\documentclass[acmsmall]{acmart}
\usepackage{ragged2e}
\AtBeginDocument{%
	}

\usepackage{multirow}
\usepackage{titlesec}
\titleformat{\subsubsection}[block] {\normalfont\bfseries}{\thesubsubsection}{1em}{}  

\begin{document}
	\title{}
	
	\begin{center}
		\textbf{\Large Domain-Specific Machine Translation to Translate Medicine Brochures in English to Sorani Kurdish}\\[1em]
		\textbf{Mariam Shamal and Hossein Hassani}\\
		University of Kurdistan Hewlêr\\
		Kurdistan Region - Iraq\\
		\texttt{mariam.shamal@ukh.edu.krd, hosseinh@ukh.edu.krd}
	\end{center}
	
	\begin{abstract}
		\begin{center}
			\textbf{\large Abstract} 
		\end{center}
		\justifying
		Access to Kurdish medicine brochures is limited, depriving Kurdish-speaking communities of critical health information. To address this problem, we developed a specialized Machine Translation (MT) model to translate English medicine brochures into Sorani Kurdish using a parallel corpus of 22,940 aligned sentence pairs from 319 brochures, sourced from two pharmaceutical companies in the Kurdistan Region of Iraq (KRI). We trained a Statistical Machine Translation (SMT) model using the Moses toolkit, conducting seven experiments that resulted in BLEU scores ranging from 22.65 to 48.93. We translated three new brochures to improve the evaluation process and encountered unknown words. We addressed unknown words through post-processing with a medical dictionary, resulting in BLEU scores of 56.87, 31.05, and 40.01. Human evaluation by native Kurdish-speaking pharmacists, physicians, and medicine users showed that 50\% of professionals found the translations consistent, while 83.3\% rated them accurate. Among users, 66.7\% considered the translations clear and felt confident using the medications. 
	\end{abstract}
	\settopmatter{printfolios=true}
	\maketitle
	
	\section{Introduction}
	Machine Translation (MT) has become essential for overcoming language barriers across various domains. In the healthcare sector, accurate and accessible communication of medical information is particularly critical, as it directly impacts patient safety and well-being. One of the main sources of this information is Consumer Medicine Information (CMI) brochures, which provide essential details about medications, including their names, ingredients, dosage instructions, potential side effects, and storage guidelines \cite{Goods}. These brochures are vital in empowering patients to safely and effectively use their medications.
	
	However, a significant challenge arises from the lack of availability of these brochures in Kurdish, which hinders access for Kurdish-speaking patients. This language gap often leads to misunderstandings, increasing the risk of medication errors and other health-related complications. Addressing this issue requires a tailored approach to machine translation that can handle the unique requirements of medical texts.
	
	Domain-specific Machine Translation offers a promising solution to accommodate specific fields' specialized vocabulary and terminologies, including healthcare. Nevertheless, translating medical content remains challenging due to the complexity of medical language and its reliance on precise terminology \cite{Ehab}. Errors in translating drug-related information, for instance, can result in incorrect dosage recommendations or misunderstandings about drug usage, posing serious risks to patient safety \cite{wolk}. Therefore, the development of accurate and specialized MT models for the medical domain is crucial to ensure that patients receive clear, precise, and culturally appropriate medication information.
	
	This research addresses the gap in Kurdish medical resources by developing a specialized MT model for translating English medicine brochures into Sorani Kurdish. To achieve this, we develop a parallel corpus of aligned English-Kurdish sentence pairs that serve as a foundation for training the model. Using the Moses toolkit, we train a Statistical Machine Translation (SMT) model and conduct various experiments to evaluate its effectiveness. The evaluation process involves quantitative metrics, such as BLEU scores, and qualitative human assessments, incorporating feedback from Kurdish-speaking medical professionals and medicine users.
	
	The rest of the paper is organized as follows: Section 2 discusses the challenges in Kurdish language text processing. Section 3 presents the historical evolution of medical translation. Section 4 reviews related work in the field. Section 5 details the methods employed to develop the domain-specific machine translation, along with the experiments conducted and their results. Finally, Section 6 concludes the paper and highlights potential directions for future work.

	\section{Challenges in Kurdish Language Text Processing}
	The Kurdish language, spoken by over 30 million people, faces significant challenges in natural language processing (NLP) due to limited resources \cite{Badawi}. These challenges arise from factors such as the existence of multiple dialects and sub-dialects, the usage of several scripts, and a lack of standardization \cite{SAb}. Additionally, there is insufficient investment in tools, parallel corpora, and online resources for Kurdish language processing. Despite efforts towards standardization, such as the Unified Kurdish Alphabet Yekgirtú, the language remains fragmented, making it difficult for readers to fully understand written Kurdish \cite{sinaa}.

	\section{Historical Evolution of Medical Translation}
	
	Medical translation is one of the oldest forms of scientific translation, closely tied to the shared history of medicine across different cultures. The vocabulary in this field largely comes from Greco-Latin roots, which has eased some linguistic challenges \cite{fishbach}. Evidence of medical translation can be traced back to ancient civilizations like Mesopotamia, where the Sumerian pharmacological lexicon preserved early medical knowledge. In ancient Greece, the Corpus Hippocraticum was instrumental in spreading medical knowledge, being translated into various languages, and laying the groundwork for modern medical texts \cite{montalt}. 
	
	Following the fall of the Western Roman Empire, much of the medical knowledge was safeguarded in the Middle East through the Graeco-Arabic translation movement. Arabic scholars, including Avicenna, translated Greek medical texts, which later made their way back to Europe through Arabic-Latin translations, aiding in the revival of European medicine and the establishment of medical education at universities like Salerno and Montpellier \cite{tracey}.
	
	Although Greek and Latin dominated scientific writing for centuries, English emerged as the primary language in the 20th century, largely due to the impact of English-speaking researchers (Karwacka, 2015). Medical translation continues to be vital for sharing knowledge despite linguistic and cultural differences and significantly impacts modern medicine \cite{Daiber}.
	
	\section{Related Work}
	This section reviews previous research relevant to the study, focusing on three areas: Domain-specific Machine Translation, Medicine-Related Machine Translation, and Translation into Kurdish Language. 
	
	\subsection{Domain-Specific Machine Translation}
	Machine Translation (MT) research has explored various approaches to address linguistic challenges, especially for morphologically rich and structurally diverse languages. \citet{Ahsan} tackled the complexities of English-to-Hindi MT by integrating rule-based (RBMT) and statistical (SMT) methods to manage Hindi’s rich morphology and flexible word order. Using an 8,768-sentence tourism domain corpus from the ICON-2008 NLP tools contest, they compared baseline SMT models, reordering rule-based models, and a hybrid Automatic Post-editing approach that used RBMT outputs as SMT inputs. Evaluation through BLEU scores and manual assessments showed significant improvements, with the hybrid methods achieving the highest BLEU score of 11.44 and reducing the Subjective Sentence Error Rate (SSER) to 68.2. This highlights the effectiveness of hybrid approaches in overcoming linguistic complexities.
	
	\citet{Rubino} explore the integration of PBMT using the MOSES toolkit and Google Translate for English-French translation, focusing on domain adaptation in the medical field to address limited parallel data. The training data consisted of 1.8 million lines from the Europarl corpus, 12 million from the United Nations corpus, and monolingual data from the News Commentary and Shuffled News Crawl corpora. The evaluation involved both human and automatic assessments, with the BLEU scores being 47.9\% for naive post-editing and 53.5\% for Oracle and the human evaluation showed a recall of 79.5\% and a precision of 40.1\% for classifying sentences for post-editing. The study demonstrated that the proposed SPE approach significantly improved translation quality, especially when combined with a classification step to select sentences that would benefit from post-editing, enhancing the translation of specialized medical terminology and linguistic structures.
	
	\subsection{Medicine-Related Machine Translation}
	\citet{Renato} investigate the use of Statistical Machine Translation (SMT) for translating medical terminologies from Spanish to Brazilian Portuguese. They developed and trained an SMT system using domain-specific parallel corpora to assess its performance against general-purpose translation systems. Validation methods included comparing translations to domain-specific references and verifying a list of 1,000 terms with the help of a native Portuguese speaker. The study used approximately 180,000 documents, containing 215,419 medical terms, with 25,038 sourced from DeCS Health Science Descriptors. Evaluation metrics like BLEU, METEOR, and TER were employed, showing that M-SMT outperformed Bing and Google in translation quality and accuracy.
	
	\citet{skians} evaluate machine translation methods for translating English medical terms, specifically ICD-11 and ICF, into French using tools like MOSES and fairseq. The study tested both statistical and neural machine translation methods on a large dataset of 499,885 sentences, with 123,445 sentences for ICD-11, 5,920 for ICF, and 24,242 validated sentences for ICD-11. The results, measured using BLEU, SacreBLEU, METEOR, and TER, showed that traditional SMT performed better for shorter sentences, while NMT was more effective for longer ones. The best model achieved 65.59 SacreBLEU, 57.50 BLEU, 46.20 METEOR, and 28.62 TER, demonstrating strong performance in translating medical terminologies.
	
	\subsection{Translation into Kurdish Language}
	
	\cite{HHSA} developed a parallel corpora for Kurdish languages, specifically Sorani and Kurmanji dialects. The corpora contain 12,327 Sorani-Kurmanji, 1,797 Kurmanji-English, and 650 Sorani-English translation pairs.
	
	\citet{masoud} developed a neural machine translation system for Sorani Kurdish, addressing challenges from limited parallel corpora. They utilized datasets like the Tanzil Corpus (92,354 sentences), TED Corpus (2,358 sentences), and KurdNet (4,663 definitions), applying tokenization methods such as WordPiece, byte-pair encoding, and NLTK’s WordPunct tokenizer. Experiments with PyTorch’s OpenNMT tested two LSTM-based configurations, integrating pre-trained embeddings (FastText for Kurdish, GloVe for English). Translation quality was evaluated using BLEU, METEOR, and TER, with datasets divided into training, validation, and testing sets. The study highlights the need for more Kurdish language resources and tools to address data imbalances and advance translation quality.
	
	\citet{amini} introduced Awta, the first large-scale parallel corpus for Sorani (Central) Kurdish-English translation, consisting of 229,222 manually aligned translation pairs across various text genres. Of these, 100,000 pairs are publicly available to promote further research. The dataset is split into 90\% for training and 10\% for testing. Evaluation using the BLEU metric revealed limitations in translation quality assessment due to the agglutinative nature of Kurdish and its rigid word-matching requirements. Neural machine translation models achieved BLEU scores of 22.72 for Kurdish-to-English and 16.81 for English-to-Kurdish. The study highlights challenges in Kurdish machine translation, such as low data visibility and the language’s morphological complexity, emphasizing the importance of parallel corpora in advancing machine translation methods.
	
	\section{Methodology}
	In this section, the methods for data collection, preprocessing, model training, and evaluation of the domain-specific machine translation model are outlined, along with the results obtained at each stage of the process.
	
	\subsection{Data Collection}
	To develop a domain-specific machine translation model for translating medical brochures from English to Sorani Kurdish, we collected 774 bilingual brochures from two local pharmaceutical companies: 166 brochures from Pioneer Pharmaceutical Company in Sulaymaniyah and 608 brochures from Awamedica Pharmaceutical Company in Erbil. These brochures, available in AI and PDF formats, included English, Arabic, and Kurdish translations with comprehensive information on medications, such as their therapeutic uses, composition, dosage, safety tips, side effects, and manufacturer details, serving as a foundational resource for the model. Figures 1 and 2 present the same sample brochures, with Figure 1 in English and Figure 2 in Sorani Kurdish.
	
	\begin{figure}
		\centering
		\fbox{%
			\includegraphics[width=0.8\linewidth]{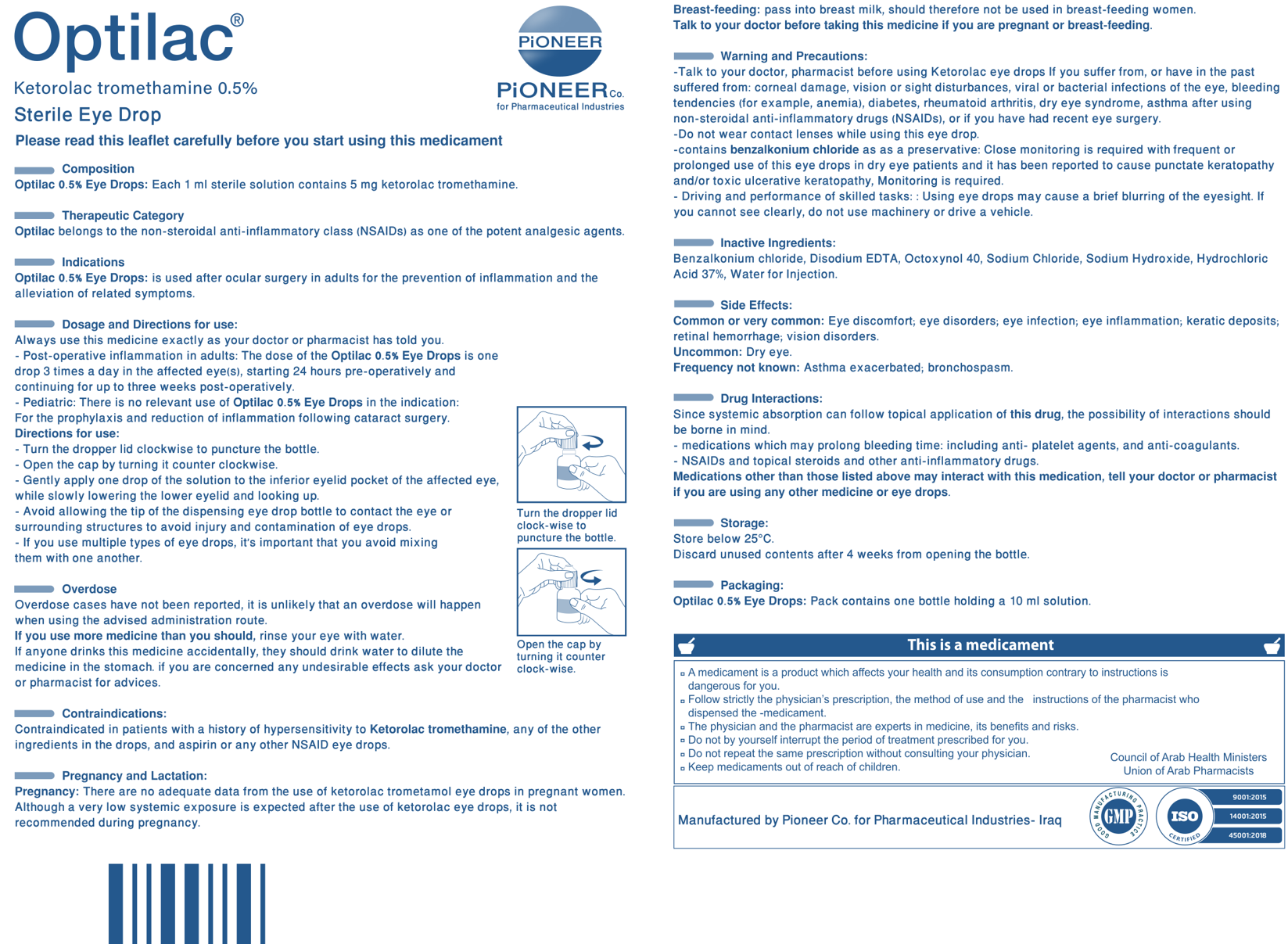}
		}
		\caption{Sample Brochure in English}
		\label{fig:brochure-english}
	\end{figure}
	
	\begin{figure}
		\centering
		\fbox{%
			\includegraphics[width=0.8\linewidth]{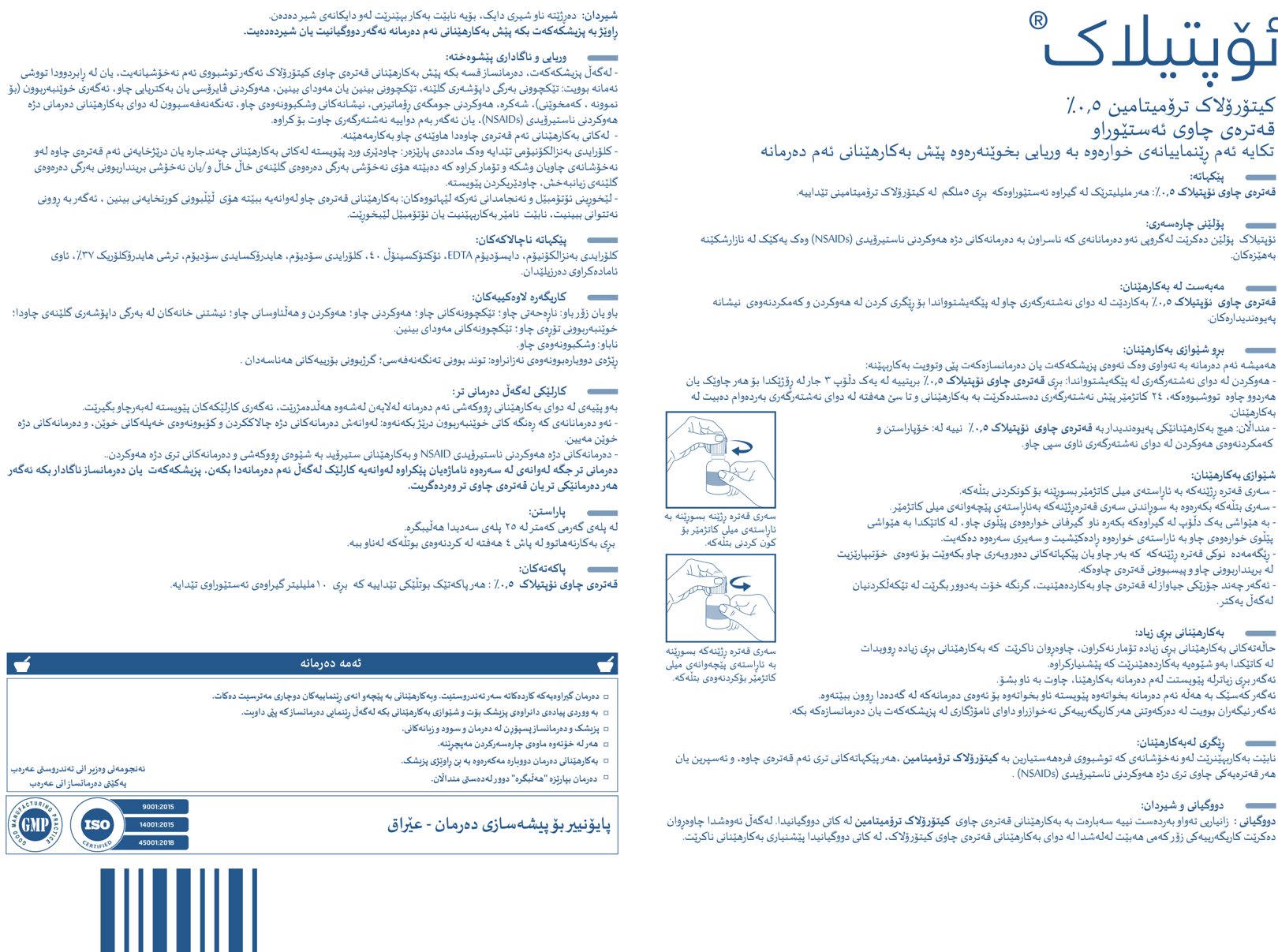}
		}
		\caption{Sample Brochure in Sorani Kurdish}
		\label{fig:brochure-kurdish}
	\end{figure}
	
	\subsection{Preprocessing}
	In this section, we outline the steps taken to prepare the dataset for our machine translation model, including data cleaning, text extraction, sentence alignment, and corpus preparation.

	\subsubsection{Data Cleaning} 
	
	We began the preprocessing phase with data cleaning, focusing on removing redundant and incomplete brochures. Incomplete brochures are defined as those lacking the Sorani-Kurdish translation. From Awamedica's collection, we removed 412 duplicates and 28 incomplete brochures, while from Pioneer's collection, we eliminated 14 duplicates and 1 incomplete brochure.
	
	Due to multiple versions of the same brochure existing, varying by batch number or approval status, we conducted the cleaning process manually. Each brochure was reviewed, and we consulted with representatives from both companies to ensure the selection of the most current versions.
	
	As a result, we excluded 426 duplicates and 29 incomplete brochures, reducing the dataset to 319 unique brochures: 168 from Awamedica and 151 from Pioneer, as shown in Table 1.
	
	\begin{table}
		\centering
		\caption{Overview of Collected Data Before and After Cleaning Process} 
		\begin{tabular}{|p{2.5cm}|p{1.7cm}|p{1.7cm}|p{1.7cm}|p{2.6cm}|} 
			\hline
			\textbf{Pharmaceutical Company} & 
			\textbf{Collected Brochures } & 
			\textbf{Redundant Brochures} & 
			\textbf{Incomplete Brochures} & 
			\textbf{Remaining Brochures After Cleaning} \\ \hline
			Awamedica & 608  & 412  & 28  & 168 \\ \hline
			Pioneer & 166  & 14  & 1  & 151 \\ \hline
			Total Brochures & 774  & 426  & 29 & 319 \\ \hline
		\end{tabular}
	\end{table}

	\subsubsection{Text Extraction}
	We utilized free online Optical Character Recognition (OCR) tools, including PDF2Go, i2OCR, Nanonets OCR, and OCR2edit, to extract English and Kurdish text from PDF and AI files. The effectiveness of these tools varied, with some performing better on specific brochures. We tested all tools to achieve optimal results, occasionally switching between them and using screenshots converted to PNG files to enhance the OCR output.
	
	The extracted text was saved in two plain text documents per brochure, one for English and one for Kurdish. While the tools performed well with English text, they were less accurate for Kurdish. i2OCR yielded the best results for Kurdish, but issues such as extra numbers, symbols, formatting inconsistencies, incorrect line breaks, and Arabic characters remained.
	
	To address these issues, we manually corrected the text, removing errors and adjusting formatting, including reformatting bullet points into paragraphs where necessary. We also deleted untranslated sections, such as drug interactions, which were missing in about 90\% of Awamedica’s Kurdish brochures. Each Kurdish brochure required 2 to 3 hours for correction, which is why we used the Lexilogos Sorani Kurdish Keyboard to speed up the process.
	
	Finally, the corrected translation pairs were stored in separate plain text files for each language.
	
	\subsubsection{Sentence Alignment}
	We performed sentence alignment using the InterText Editor on the Linux Ubuntu 22 OS. The InterText Editor is a tool designed for aligning parallel texts, offering both manual and automated alignment through Hunalign. It supports multiple languages and provides a variety of editing functions, including merging, splitting, and repositioning sentences \cite{intertext}.
	
	First, we loaded the English and Kurdish texts into InterText to ensure accessibility for both languages. We then used Hunalign’s automatic alignment to generate an initial alignment based on sentence structure and linguistic patterns.
	
	Following this, we conducted a manual review to correct any inaccuracies, including merging or splitting sentences to enhance alignment quality. After confirming its accuracy, we saved the aligned texts in plain text, XML, and TMX formats. Refer to Figure 3 for a sample of sentence alignment performed using the InterText Editor.
	
	\begin{figure}
		\centering
		\fbox{%
			\includegraphics[width=0.9\linewidth]{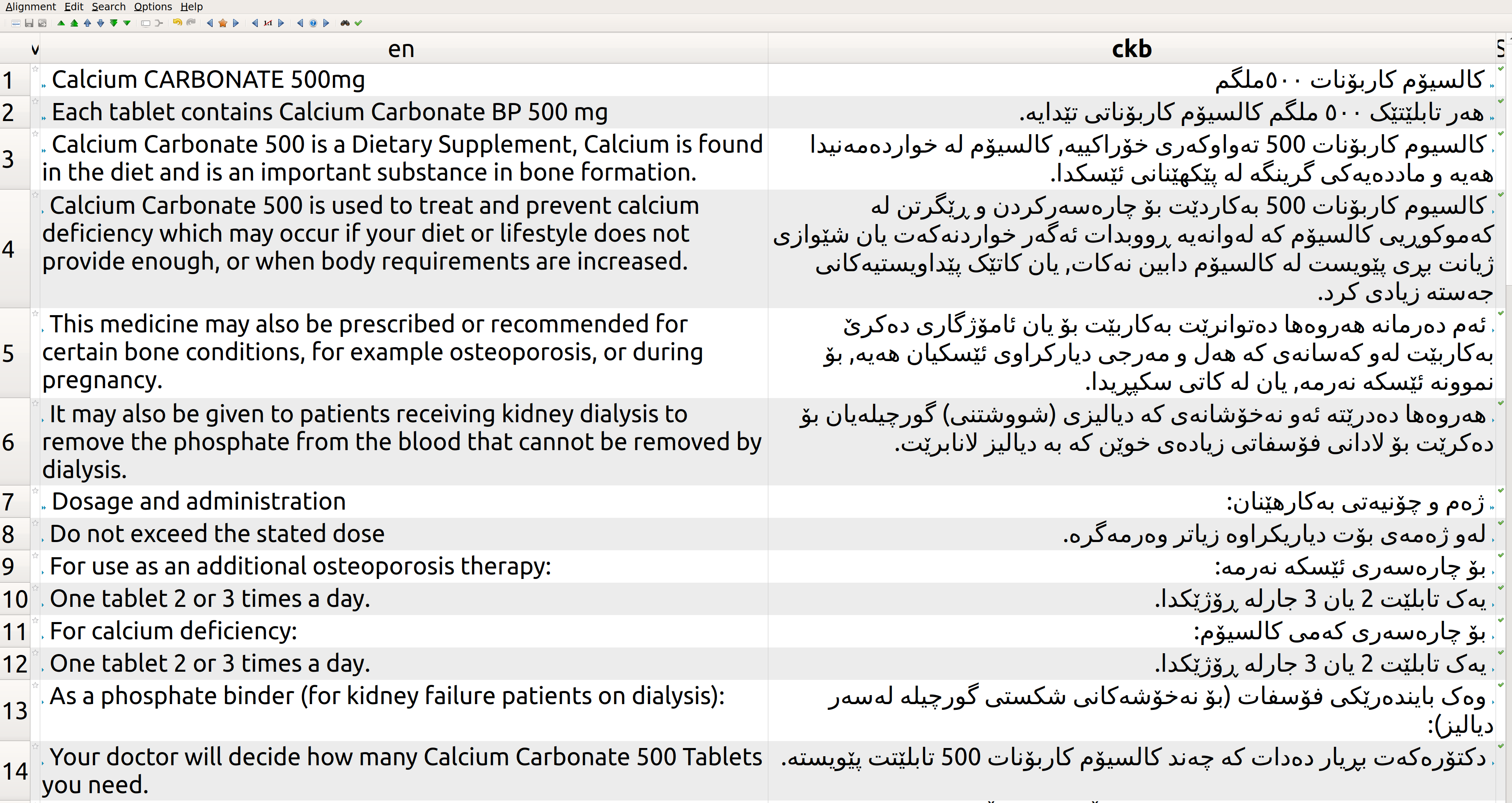}
		}
		\caption{Sentence Alignment in InterText Editor}
		\label{fig:enter-label}
	\end{figure}

	\subsubsection{Corpus Preparation}
	
	Upon finalizing the alignment process, we created several versions of our corpus for training our machine translation model. In the first version, we merged the source texts of each brochure into one text file and the target texts into another, resulting in 22,940 lines, ensuring uniform structure across both languages. 
	
	For the second version, we generated shuffled versions of both the source and target text files, ensuring that content alignment was preserved. 
	
	In the third version, we used XML format and added brochure number tags to the 319 brochure texts, raising the total line count to 23,580.
	
	For the fourth version, we mixed the brochures, selecting sentences randomly from their content while maintaining the alignment between the source and target texts. This resulted in a total of 23,899 lines.
	
	In the fifth version, category tags were added to the brochures, resulting in 76 unique categories. The brochures were organized by category, shuffled within their respective groups, and the content was further mixed by randomly selecting sentences. This version maintained the overall structure while improving diversity.
	
	In the sixth version, we applied the undersampling technique to reduce the length of the larger brochures by randomly removing sentences and aligning them with the size of the smaller brochures within each category. This process reduced the total number of lines to 16,767.
	
	Finally, for the seventh version, we used the oversampling technique to enlarge the shorter brochures by duplicating sentences, ensuring consistent length across all brochures within each category, with shuffling and random selection resulting in the total line count increasing to 32,784.

	\subsection{Experiments and The Trained Model}
	In this section, we detail the experiments conducted and the training of the machine translation model with Moses.
	
	\subsubsection{Experiments}
	To evaluate the impact of various data handling strategies on training our machine translation model, we conducted seven experiments using the different versions of our corpus. Each version was split into 90\% training and 10\% testing sets to balance data sufficiency for training and effective evaluation.
	
	In the first experiment, we used the original corpus, containing 22,940 aligned lines of source and target texts, split into 20,646 training lines and 2,294 testing lines while maintaining brochure sequence and alignment. In the second experiment, we used the shuffled second version of the corpus, dividing it similarly.
	
	For the third experiment, we utilized the XML-formatted third version of the corpus, splitting it based on brochure numbers, with 287 brochures allocated for training and 32 for testing. The fourth experiment involved a shuffled XML fourth version, dividing 20,506 training lines and 2,434 testing lines, with content shuffled at the brochure level.
	
	In the fifth experiment, we worked with the fifth version of the corpus, where we grouped brochures by category, shuffled them, and split the data into 20,612 training lines and 2,328 testing lines. For the sixth experiment, we applied the undersampling version, reducing brochure lengths to the smallest line count within each category, resulting in a balanced dataset of 16,767 lines, with 15,056 for training and 1,711 for testing.
	
	Finally, in the seventh experiment, we employed the oversampling version to equalize brochure lengths to the maximum line count within each category, increasing the corpus size to 32,784 lines, with 29,475 for training and 3,309 for testing.
	
	\subsubsection{Moses Training}
	We utilized the Moses statistical machine translation engine for training, which is an open-source tool with features such as linguistic theory-based factors, confusion network decoding, and simplified data structures for translation and language modeling. It supports the training, optimization, and execution of translation processes, including phrase transfer, language modeling, and decoding to generate optimal translations \cite{koehn-moses}.
	
	We began each experiment by tokenizing the corpus into individual word tokens. Following this, a truecaser was trained to gather statistical data, and a recasing procedure was carried out using a Moses script. Misaligned or empty sentences were removed to ensure data clarity. Word alignment was performed using Giza++, which generated a phrase table mapping source phrases to their target equivalents. Simultaneously, an n-gram language model for Sorani Kurdish was trained using KenLM, incorporating smoothing techniques to handle unseen word sequences and improve translation fluency. Finally, the Moses decoder was trained using the phrase table and language model to generate the translations. Figure 4 illustrates the Moses training process.
	
	\begin{figure}
		\centering
		\fbox{%
			\includegraphics[width=0.7\linewidth]{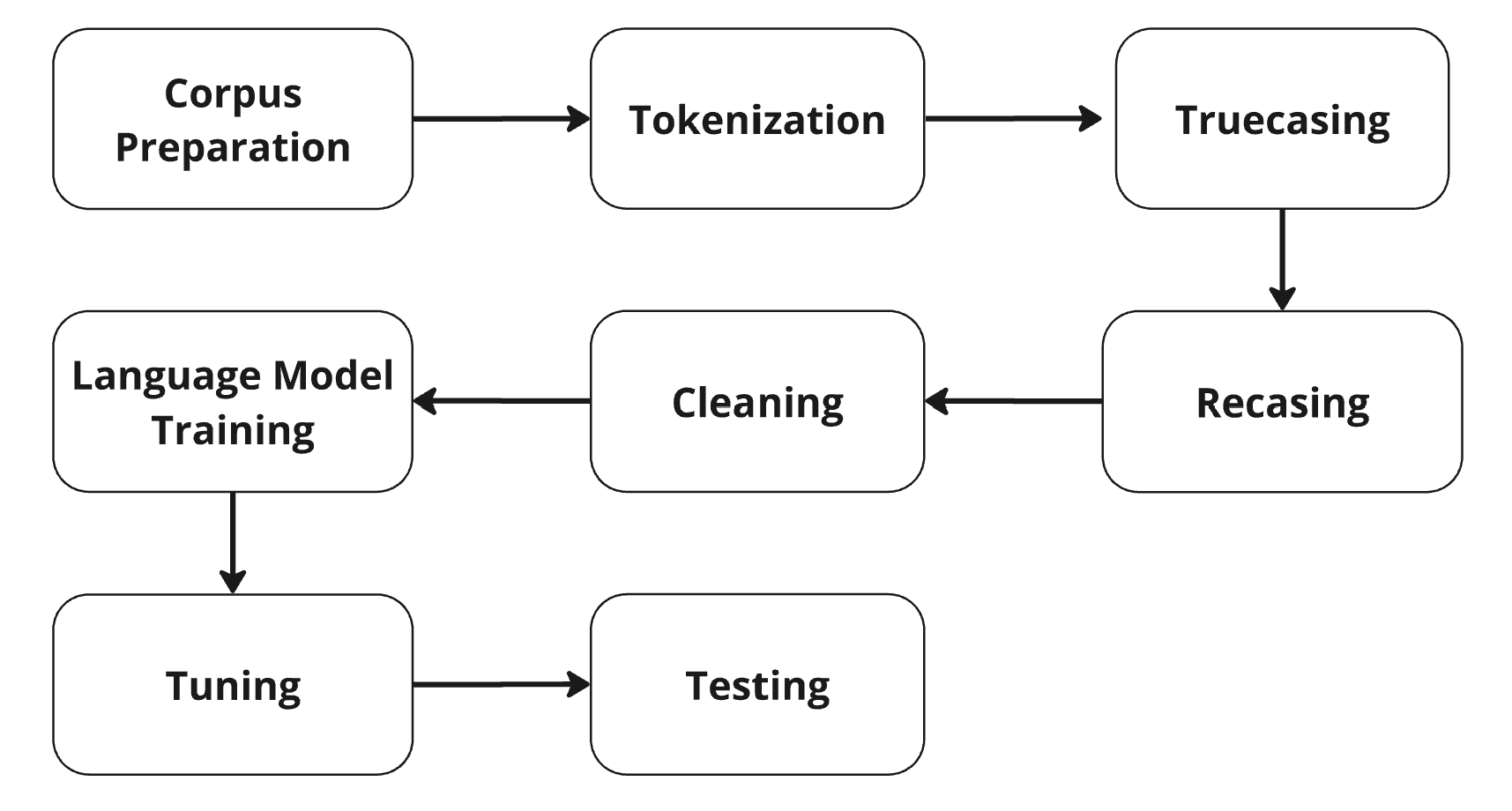}
		}
		\caption{Moses Training Overview}
		\label{fig:moses_training}
	\end{figure}
	
	\subsection{Evaluation}
	This section assesses translation quality using quantitative automatic evaluation metrics and qualitative human evaluations.
	
	\subsection{Automatic Evaluation}
	We evaluated the quality of the translations using the Bilingual Evaluation Understudy (BLEU) score. This widely used automatic metric measures the degree of n-gram overlap between machine-generated and reference translations. The BLEU score is based on precision and incorporates a brevity penalty to prevent excessively short translations. The score ranges from 0 (no overlap) to 1 (perfect match) and is often expressed as a percentage (0 to 100) for clarity, with higher scores indicating better alignment with the reference translations \cite{BLEU}.
	
	We obtained various BLEU scores from our experiments, reflecting different levels of translation quality. The first experiment, using the original aligned corpus, achieved a score of 26.73. Shuffling the text files in the second experiment raised the score to 31.34, while shuffling both brochures and content in the fourth experiment further improved it to 31.78. Grouping brochures by category in the fifth experiment resulted in a score of 31.37. The undersampling technique in the sixth experiment produced a score of 30.37, and the oversampling technique in the seventh experiment achieved the highest score of 48.93. In contrast, the third experiment, which focused solely on the number of brochures, yielded the lowest score of 22.65, indicating a less effective approach. The BLEU scores for all experiments are shown in Table 2.
	
	\begin{table}
		\centering
		\caption{BLEU Scores of the Experiments} 
		\begin{tabular}{|p{2cm}|p{2cm}|} 
			\hline
			\textbf{Experiments} & 
			\textbf{BLEU Scores} \\ \hline
			Experiment 1 & 26.73  \\ \hline
			Experiment 2 & 31.34 \\ \hline
			Experiment 3 & 22.65 \\ \hline
			Experiment 4 & 31.78\\ \hline
			Experiment 5 & 31.37  \\ \hline
			Experiment 6 & 30.37\\ \hline
			Experiment 7 & 48.93 \\ \hline
		\end{tabular}
	\end{table}
	
	\subsection{Human Evaluation}
	
	To further assess the quality of our translations, we performed a qualitative human evaluation. We translated three new English brochures using our best-performing model. These brochures were selected based on their moderate length and representation of different categories and scenarios. The first brochure, selected from our test set, contained 42 lines. The second brochure, with 50 lines, was a new input from a category previously addressed in earlier translations. The third brochure, completely new, contained 27 lines.
	
	For the two new brochures, we employed OCR to extract the text, followed by manual refinement to address inconsistencies. The initial translations revealed issues such as unknown words, incorrect translations, and structural errors, resulting in BLEU scores of 25.79, 6.46, and 12.25, respectively. To improve the translations, we performed post-processing steps, including replacing unknown words with equivalents from a medical dictionary. We created our dictionary, incorporated ambiguous terms, and consulted experts in cases with multiple possible translations. For unresolved cases, we used the Google Cloud Translation API. After these modifications, BLEU scores significantly improved to 56.87, 31.05, and 40.01. A summary of BLEU scores before and after post-editing is presented in Table 3.
	
	\begin{table}[ht!]
		\centering
		\caption{BLEU Scores of Pre- and Post-editing}
		\begin{tabular}{|l|l|l|l|} 
			\hline
			\textbf{Brochures} & \multicolumn{1}{c|}{\textbf{Before Post-Editing}} & \multicolumn{1}{c|}{\textbf{After Post-Editing}} \\ 
			\hline
			Brochure 1 & 25.79 & 56.87 \\ \hline
			Brochure 2 & 6.46  & 31.05 \\ \hline
			Brochure 3 & 12.25  & 40.01  \\ \hline
		\end{tabular}
		\label{fontsres}
	\end{table}

	We selected nine native Kurdish-speaking evaluators, comprising pharmacists, physicians, and medicine users, to assess the post-processed translations. The evaluators rated the translations on understandability, terminological accuracy, and consistency using a 4-point Likert scale via Google Forms. The results revealed that 50\% of professionals rated the translations as mostly understandable and consistent, with 83.3\% finding the terminology accurate, as shown in Figure 5. Furthermore, 66.7\% of the users found the translations mostly understandable and expressed confidence in using the medications after reading the brochures, as illustrated in Figure 6.
	
	\begin{figure}
		\centering
		\fbox{%
			\includegraphics[width=1\linewidth]{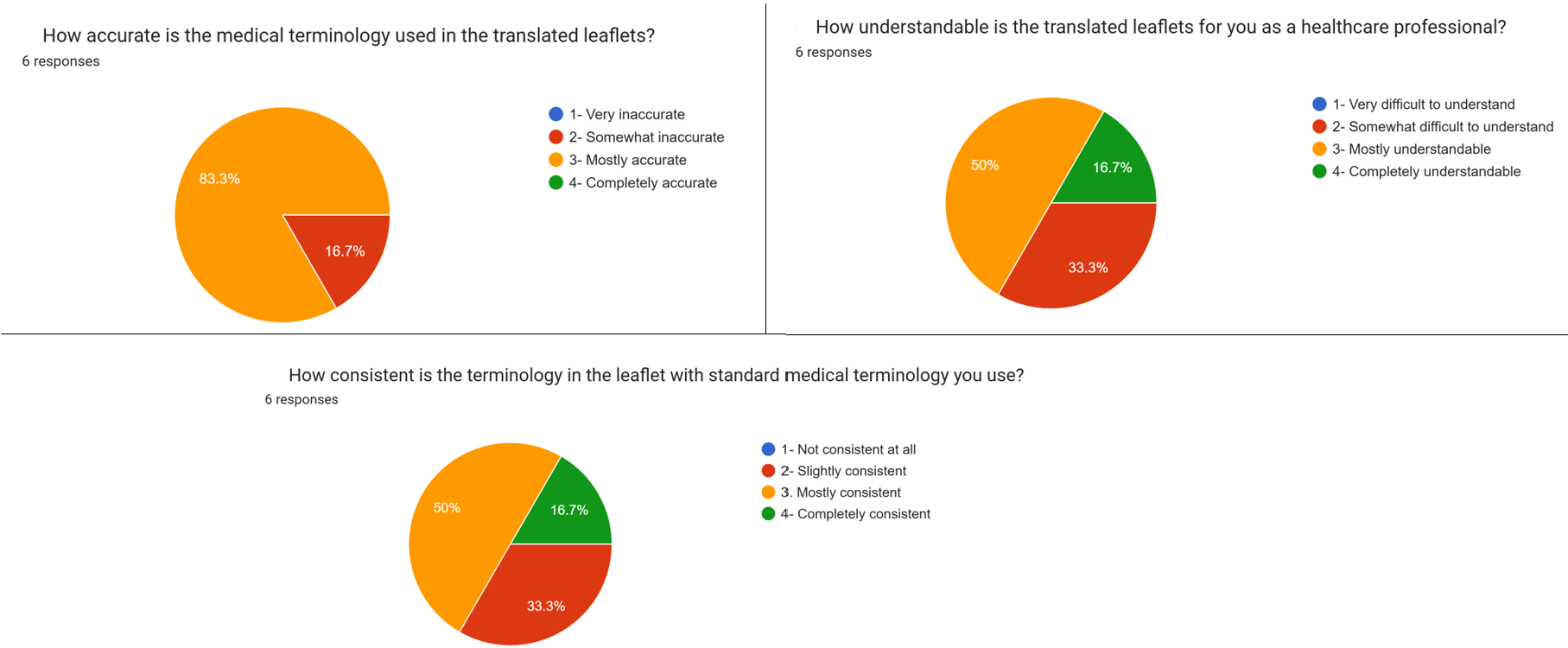}
		}
		\caption{Professionals' Feedback on the Understandability, Accuracy, and Consistency of the Translations}
		\label{fig:professional_evaluation}
	\end{figure}
	
	\begin{figure}
		\centering
		\fbox{%
			\includegraphics[width=1\linewidth]{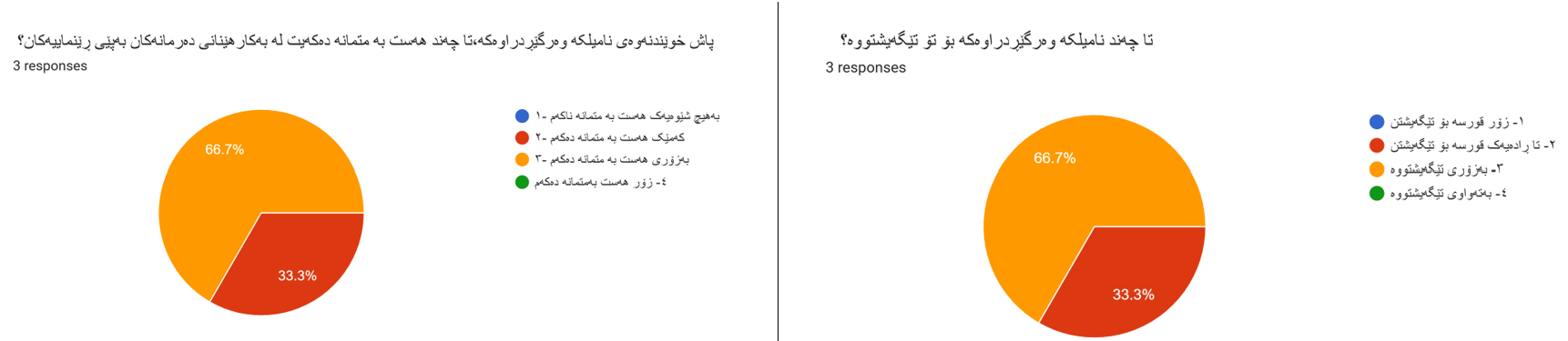}
		}
		\caption{Medicine Users' Feedback on the Translations' Understandability and Confidence in Medication Use}
		\label{fig:user_evaluation}
	\end{figure}
	
	\section{Conclusion and Future Work}
	This paper presents the development of a machine translation model for translating medicine brochures from English to Sorani Kurdish. A parallel corpus of 22,940 aligned sentence pairs from 319 brochures was created and pre-processed to correct redundancy and formatting issues. Using this corpus, a statistical machine translation model was trained with the Moses engine. The translation system was evaluated using BLEU scores, which ranged from 22.65 to 48.93, and further improved through post-processing by replacing unknown words using a medical dictionary. Post-processed translations achieved BLEU scores of 56.87, 31.05, and 40.01. Human evaluation by nine native Kurdish-speaking assessors, including pharmacists, physicians, and medicine users, showed that 50\% of professionals found the translations mostly understandable and consistent, with 83.3\% rating the terminology as accurate. Additionally, 66.7\% of users found the translations understandable and expressed confidence in using the medications.
	
	Our resource provides a solid foundation for enhancing Kurdish machine translation, offering opportunities for future improvements. To improve the system, several directions can be explored. First, gathering more data from a single local pharmaceutical company could customize the model to its unique terminology, thereby increasing translation accuracy and relevance. Automating the post-editing process with probability-based models would assist in handling unknown words and ensuring translations are more contextually appropriate. Incorporating context-aware and rule-based systems could address challenges like incorrect translations and broken sentences, leading to better overall quality.  Furthermore, leveraging state-of-the-art neural network approaches could further boost performance, especially with larger datasets. Experimenting with various data splits may optimize resource allocation and improve the model's generalization capabilities. Finally, evaluating and comparing the current system with existing translation platforms, such as Google Translator and Claude AI, could provide valuable insights into its effectiveness and potential for improvement.
	
	\begin{acks}
		We sincerely thank the Awamedica Pharmaceutical Company staff, especially Dr. Shakhawan Omar, Dr. Dilan Zakarya, and Dr. Ghada Abdulqadir, for providing the essential English and Kurdish medicine brochures. We also extend our thanks to Dr. Jaafar Haydar, Mrs. Lavin Dhyaalddin, and the team at Pioneer Pharmaceutical Company for their invaluable assistance in collecting the medicine brochures used in this study. Special thanks to Prof. Dr. Kawa Fariq, Prof. Dr. Salah Abubakr, Dr. Narmeen Jamal, Dr. Dilman Azad, Dr. Valy Suliman, and Dr. Helin Haydar for their insightful feedback on the model. We also appreciate Mrs. Rowshan Jalal, Mr. Sherwan Jalal, and Liza Suliman for their evaluations as medicine users.
	\end{acks}
	
	\bibliographystyle{ACM-Reference-Format}
	
	\bibliography{DomainSpecificMT-Kurdish}


\begin{thebibliography}{20}


\ifx \showCODEN    \undefined \def \showCODEN     #1{\unskip}     \fi
\ifx \showDOI      \undefined \def \showDOI       #1{#1}\fi
\ifx \showISBNx    \undefined \def \showISBNx     #1{\unskip}     \fi
\ifx \showISBNxiii \undefined \def \showISBNxiii  #1{\unskip}     \fi
\ifx \showISSN     \undefined \def \showISSN      #1{\unskip}     \fi
\ifx \showLCCN     \undefined \def \showLCCN      #1{\unskip}     \fi
\ifx \shownote     \undefined \def \shownote      #1{#1}          \fi
\ifx \showarticletitle \undefined \def \showarticletitle #1{#1}   \fi
\ifx \showURL      \undefined \def \showURL       {\relax}        \fi
\providecommand\bibfield[2]{#2}
\providecommand\bibinfo[2]{#2}
\providecommand\natexlab[1]{#1}
\providecommand\showeprint[2][]{arXiv:#2}

\bibitem[Ahmadi(2019a)]%
        {sinaa}
\bibfield{author}{\bibinfo{person}{Sina Ahmadi}.}
  \bibinfo{year}{2019}\natexlab{a}.
\newblock \showarticletitle{{A Rule-Based Kurdish Text Transliteration
  System}}.
\newblock \bibinfo{journal}{\emph{ACM Transactions on Asian and Low-Resource
  Language Information Processing}}  \bibinfo{volume}{18} (\bibinfo{date}{01}
  \bibinfo{year}{2019}), \bibinfo{pages}{1--8}.
\newblock
\urldef\tempurl%
\url{https://doi.org/10.1145/3278623}
\showDOI{\tempurl}


\bibitem[Ahmadi(2019b)]%
        {SAb}
\bibfield{author}{\bibinfo{person}{Sina Ahmadi}.}
  \bibinfo{year}{2019}\natexlab{b}.
\newblock \bibinfo{title}{{Why Does Kurdish Language Processing Matter?}}
\newblock
\newblock
\urldef\tempurl%
\url{https://sinaahmadi.github.io/posts/why-kurdish-language-processing-matters.html/(Accessed_on_15-01-2020)}
\showURL{%
\tempurl}


\bibitem[Ahmadi et~al\mbox{.}(2020)]%
        {HHSA}
\bibfield{author}{\bibinfo{person}{Sina Ahmadi}, \bibinfo{person}{Hossein
  Hassani}, {and} \bibinfo{person}{Daban Jaff}.}
  \bibinfo{year}{2020}\natexlab{}.
\newblock \showarticletitle{{Leveraging multilingual news websites for building
  a Kurdish parallel corpus}}.
\newblock \bibinfo{journal}{\emph{ACM Transactions on Asian and Low-Resource
  Language Information Processing}} (\bibinfo{date}{10} \bibinfo{year}{2020}).
\newblock
\newblock
\shownote{Available at: \url{https://dl.acm.org/doi/10.1145/3511806}}.


\bibitem[Ahmadi and Masoud(2020)]%
        {masoud}
\bibfield{author}{\bibinfo{person}{Sina Ahmadi} {and} \bibinfo{person}{Maraim
  Masoud}.} \bibinfo{year}{2020}\natexlab{}.
\newblock \showarticletitle{{Towards Machine Translation for the Kurdish
  Language}}.
\newblock  (\bibinfo{date}{10} \bibinfo{year}{2020}).
\newblock


\bibitem[Ahsan et~al\mbox{.}(2010)]%
        {Ahsan}
\bibfield{author}{\bibinfo{person}{Arafat Ahsan}, \bibinfo{person}{Prasanth
  Kolachina}, \bibinfo{person}{Sudheer Kolachina}, \bibinfo{person}{Dipti
  Sharma}, {and} \bibinfo{person}{Rajeev Sangal}.}
  \bibinfo{year}{2010}\natexlab{}.
\newblock \showarticletitle{{Coupling Statistical Machine Translation with
  Rule-based Transfer and Generation}}.
\newblock \bibinfo{journal}{\emph{AMTA 2010 - 9th Conference of the Association
  for Machine Translation in the Americas}} (\bibinfo{date}{01}
  \bibinfo{year}{2010}).
\newblock


\bibitem[Amini et~al\mbox{.}(2021)]%
        {amini}
\bibfield{author}{\bibinfo{person}{Zhila Amini}, \bibinfo{person}{Aran
  Emînî}, \bibinfo{person}{Hawre Hosseini}, \bibinfo{person}{Mehran
  Mansouri}, {and} \bibinfo{person}{Daban Jaff}.}
  \bibinfo{year}{2021}\natexlab{}.
\newblock \showarticletitle{CENTRAL KURDISH MACHINE TRANSLATION: FIRST LARGE
  SCALE PARALLEL CORPUS AND EXPERIMENTS}.
\newblock  (\bibinfo{date}{06} \bibinfo{year}{2021}).
\newblock


\bibitem[Badawi(2023)]%
        {Badawi}
\bibfield{author}{\bibinfo{person}{Soran Badawi}.}
  \bibinfo{year}{2023}\natexlab{}.
\newblock \showarticletitle{Using Multilingual Bidirectional Encoder
  Representations from Transformers on Medical Corpus for Kurdish Text
  Classification}.
\newblock \bibinfo{journal}{\emph{ARO-THE SCIENTIFIC JOURNAL OF KOYA
  UNIVERSITY}}  \bibinfo{volume}{11} (\bibinfo{date}{01} \bibinfo{year}{2023}),
  \bibinfo{pages}{10--15}.
\newblock
\urldef\tempurl%
\url{https://doi.org/10.14500/aro.11088}
\showDOI{\tempurl}


\bibitem[Daiber and Daiber(2021)]%
        {Daiber}
\bibfield{author}{\bibinfo{person}{Hans Daiber} {and} \bibinfo{person}{Helga
  Daiber}.} \bibinfo{year}{2021}\natexlab{}.
\newblock \bibinfo{booktitle}{\emph{From the greeks to the Arabs and beyond.
  volume I, Graeco-Syriaca and Arabica}}.
\newblock \bibinfo{publisher}{Brill}.
\newblock


\bibitem[Ehab et~al\mbox{.}(2018)]%
        {Ehab}
\bibfield{author}{\bibinfo{person}{Rana Ehab}, \bibinfo{person}{Eslam Amer},
  {and} \bibinfo{person}{Mahmoud Gadallah}.} \bibinfo{year}{2018}\natexlab{}.
\newblock \showarticletitle{{Example-Based English to Arabic Machine
  Translation: Matching Stage Using Internal Medicine Publications}}.
\newblock
\urldef\tempurl%
\url{https://doi.org/10.1145/3220267.3220294}
\showDOI{\tempurl}


\bibitem[Fischbach(1986)]%
        {fishbach}
\bibfield{author}{\bibinfo{person}{Henry Fischbach}.}
  \bibinfo{year}{1986}\natexlab{}.
\newblock \showarticletitle{{Some Anatomical and Physiological Aspects of
  Medical Translation: Lexical equivalence, ubiquitous references and
  universality of subject minimize misunderstanding and maximize transfer of
  meaning}}.
\newblock \bibinfo{journal}{\emph{Meta}} \bibinfo{volume}{31},
  \bibinfo{number}{1} (\bibinfo{year}{1986}), \bibinfo{pages}{16--21}.
\newblock
\urldef\tempurl%
\url{https://doi.org/10.7202/002743ar}
\showDOI{\tempurl}


\bibitem[{Goods}(2020)]%
        {Goods}
\bibfield{author}{\bibinfo{person}{{Goods}}.} \bibinfo{year}{2020}\natexlab{}.
\newblock \bibinfo{title}{{Consumer Medicines Information Therapeutic Goods
  Administration}}.
\newblock
\newblock
\urldef\tempurl%
\url{https://www.tga.gov.au/products/australian-register-therapeutic-goods-artg/consumer-medicines-information-cmi}
\showURL{%
\tempurl}


\bibitem[Koehn et~al\mbox{.}(2007)]%
        {koehn-moses}
\bibfield{author}{\bibinfo{person}{Philipp Koehn}, \bibinfo{person}{Hieu
  Hoang}, \bibinfo{person}{Alexandra Birch}, \bibinfo{person}{Chris
  Callison-Burch}, \bibinfo{person}{Marcello Federico}, \bibinfo{person}{Nicola
  Bertoldi}, \bibinfo{person}{Brooke Cowan}, \bibinfo{person}{Wade Shen},
  \bibinfo{person}{Christine Moran}, \bibinfo{person}{Richard Zens},
  \bibinfo{person}{Chris Dyer}, \bibinfo{person}{Ond{\v{r}}ej Bojar},
  \bibinfo{person}{Alexandra Constantin}, {and} \bibinfo{person}{Evan Herbst}.}
  \bibinfo{year}{2007}\natexlab{}.
\newblock \showarticletitle{{Moses: Open Source Toolkit for Statistical Machine
  Translation}}. In \bibinfo{booktitle}{\emph{Proceedings of the 45th Annual
  Meeting of the Association for Computational Linguistics Companion Volume
  Proceedings of the Demo and Poster Sessions}},
  \bibfield{editor}{\bibinfo{person}{Sophia Ananiadou}} (Ed.).
  \bibinfo{publisher}{Association for Computational Linguistics},
  \bibinfo{address}{Prague, Czech Republic}, \bibinfo{pages}{177--180}.
\newblock
\urldef\tempurl%
\url{https://aclanthology.org/P07-2045}
\showURL{%
\tempurl}


\bibitem[Marie et~al\mbox{.}(2021)]%
        {Rubino}
\bibfield{author}{\bibinfo{person}{Benjamin Marie}, \bibinfo{person}{Atsushi
  Fujita}, {and} \bibinfo{person}{Raphael Rubino}.}
  \bibinfo{year}{2021}\natexlab{}.
\newblock \showarticletitle{{Scientific Credibility of Machine Translation
  Research: A Meta-Evaluation of 769 Papers}}. In
  \bibinfo{booktitle}{\emph{Proceedings of the 59th Annual Meeting of the
  Association for Computational Linguistics and the 11th International Joint
  Conference on Natural Language Processing (Volume 1: Long Papers)}},
  \bibfield{editor}{\bibinfo{person}{Chengqing Zong}, \bibinfo{person}{Fei
  Xia}, \bibinfo{person}{Wenjie Li}, {and} \bibinfo{person}{Roberto Navigli}}
  (Eds.). \bibinfo{publisher}{Association for Computational Linguistics},
  \bibinfo{address}{Online}, \bibinfo{pages}{7297--7306}.
\newblock
\urldef\tempurl%
\url{https://doi.org/10.18653/v1/2021.acl-long.566}
\showDOI{\tempurl}


\bibitem[Montalt et~al\mbox{.}(2018)]%
        {montalt}
\bibfield{author}{\bibinfo{person}{Vicent Montalt},
  \bibinfo{person}{Karen~Korning Zethsen}, {and} \bibinfo{person}{Wioleta
  Karwacka}.} \bibinfo{year}{2018}\natexlab{}.
\newblock \showarticletitle{{Medical translation in the 21st century –
  challenges and trends}}.
\newblock \bibinfo{journal}{\emph{MonTI. Monographs in translation and
  interpreting}} \bibinfo{number}{10} (\bibinfo{date}{Dec.}
  \bibinfo{year}{2018}), \bibinfo{pages}{27–42}.
\newblock
\urldef\tempurl%
\url{https://www.e-revistes.uji.es/index.php/monti/article/view/3684}
\showURL{%
\tempurl}


\bibitem[Papineni et~al\mbox{.}(2002)]%
        {BLEU}
\bibfield{author}{\bibinfo{person}{Kishore Papineni}, \bibinfo{person}{Salim
  Roukos}, \bibinfo{person}{Todd Ward}, {and} \bibinfo{person}{Wei~Jing Zhu}.}
  \bibinfo{year}{2002}\natexlab{}.
\newblock \showarticletitle{{BLEU: a Method for Automatic Evaluation of Machine
  Translation}}.
\newblock  (\bibinfo{date}{10} \bibinfo{year}{2002}).
\newblock
\urldef\tempurl%
\url{https://doi.org/10.3115/1073083.1073135}
\showDOI{\tempurl}


\bibitem[Renato et~al\mbox{.}(2018)]%
        {Renato}
\bibfield{author}{\bibinfo{person}{Alejandro Renato}, \bibinfo{person}{José
  Castaño}, \bibinfo{person}{Maria Ávila Williams}, \bibinfo{person}{Hernan
  Berinsky}, \bibinfo{person}{Maria Gambarte}, \bibinfo{person}{Hee Park},
  \bibinfo{person}{David Perez}, \bibinfo{person}{Carlos Otero}, {and}
  \bibinfo{person}{Daniel Luna}.} \bibinfo{year}{2018}\natexlab{}.
\newblock \showarticletitle{{A Machine Translation Approach for Medical
  Terms}}. \bibinfo{pages}{369--378}.
\newblock
\urldef\tempurl%
\url{https://doi.org/10.5220/0006555003690378}
\showDOI{\tempurl}


\bibitem[Skianis et~al\mbox{.}(2020)]%
        {skians}
\bibfield{author}{\bibinfo{person}{Konstantinos Skianis}, \bibinfo{person}{Yann
  Briand}, {and} \bibinfo{person}{Florent Desgrippes}.}
  \bibinfo{year}{2020}\natexlab{}.
\newblock \showarticletitle{{Evaluation of Machine Translation Methods Applied
  to Medical Terminologies}}. \bibinfo{pages}{59--69}.
\newblock
\urldef\tempurl%
\url{https://doi.org/10.18653/v1/2020.louhi-1.7}
\showDOI{\tempurl}


\bibitem[Tracey(2018)]%
        {tracey}
\bibfield{author}{\bibinfo{person}{David Tracey}.}
  \bibinfo{year}{2018}\natexlab{}.
\newblock \showarticletitle{{An Early History of Medical Translation}}.
\newblock \bibinfo{journal}{\emph{JDDG: Journal der Deutschen Dermatologischen
  Gesellschaft}}  \bibinfo{volume}{16} (\bibinfo{date}{10}
  \bibinfo{year}{2018}), \bibinfo{pages}{1300--1301}.
\newblock
\urldef\tempurl%
\url{https://doi.org/10.1111/ddg.13667}
\showDOI{\tempurl}


\bibitem[Vondřička(2016)]%
        {intertext}
\bibfield{author}{\bibinfo{person}{Pavel Vondřička}.}
  \bibinfo{year}{2016}\natexlab{}.
\newblock \bibinfo{booktitle}{\emph{{InterText Editor v1.5 Comprehensive
  Guide}}}.
\newblock
\urldef\tempurl%
\url{http://wanthalf.saga.cz/intertext}
\showURL{%
\tempurl}


\bibitem[Wołk and Marasek(2014)]%
        {wolk}
\bibfield{author}{\bibinfo{person}{Krzysztof Wołk} {and}
  \bibinfo{person}{Krzysztof Marasek}.} \bibinfo{year}{2014}\natexlab{}.
\newblock \showarticletitle{{Polish -English Statistical Machine Translation of
  Medical Texts}}.
\newblock \bibinfo{journal}{\emph{New Research in Multimedia and Internet
  Systems, Springer}}  \bibinfo{volume}{314} (\bibinfo{date}{09}
  \bibinfo{year}{2014}).
\newblock
\showISBNx{978-3-319-10382-2}
\urldef\tempurl%
\url{https://doi.org/10.1007/978-3-319-10383-9_16}
\showDOI{\tempurl}


\end{thebibliography}
		
\end{document}